\documentclass{article}

\usepackage{microtype}
\usepackage{multirow}
\usepackage{graphicx}
\usepackage{subfigure}
\usepackage{booktabs} 
\usepackage[super]{nth}
\usepackage{hyperref}

\usepackage[accepted]{arxiv}

\usepackage[T1]{fontenc}
\usepackage{tgpagella}

\usepackage{amsmath}
\usepackage{amssymb}
\usepackage{mathtools}
\usepackage{amsthm}

\usepackage{dirtytalk}
\usepackage[capitalize,noabbrev]{cleveref}

\theoremstyle{plain}
\newtheorem{theorem}{Theorem}[section]

\theoremstyle{definition}
\newtheorem{definition}[theorem]{Definition}

\theoremstyle{remark}

\usepackage[textsize=tiny]{todonotes}

\icmltitlerunning{Against Empirical AI Alignment}

\begin{document}

\twocolumn[
\icmltitle{
\LARGE{A Statistical Case Against Empirical Human--AI Alignment
}}



\icmlsetsymbol{equal}{*}

\begin{icmlauthorlist}
\icmlauthor{Julian Rodemann}{ds}
\icmlauthor{Esteban Garcés Arias}{ds,mcml}
\icmlauthor{Christoph Luther}{v,cs}
\icmlauthor{Christoph Jansen}{lanc}
\icmlauthor{Thomas Augustin}{ds}
\end{icmlauthorlist}

\icmlaffiliation{ds}{Department of Statistics, LMU Munich, Germany}
\icmlaffiliation{mcml}{Munich Center for Machine Learning (MCML), Germany}
\icmlaffiliation{v}{Research Group Neuroinformatics, Faculty of Computer Science, University of Vienna, Vienna, Austria}
\icmlaffiliation{cs}{Doctoral School Computer Science, Faculty of Computer Science, University of Vienna, Vienna, Austria}
\icmlaffiliation{lanc}{School of Computing \& Communications, Lancaster University Leipzig, Germany}

\icmlcorrespondingauthor{Julian Rodemann}{mail@julian-rodemann.de}


\vskip 0.3in
]



\printAffiliationsAndNotice{}  

\begin{abstract}


Empirical human--AI alignment aims to make AI systems act in line with observed human behavior.
While noble in its goals, we argue that empirical alignment can inadvertently introduce statistical biases that warrant caution. 
This position paper thus advocates against naive empirical alignment, offering \textit{prescriptive} alignment and \textit{a posteriori} empirical alignment as alternatives. 
We substantiate our principled argument by tangible examples like human-centric decoding of language models. 

\end{abstract}













\section{Introduction}\label{sec:intro}

\begin{figure*}
    \centering
    \includegraphics[width=0.95\linewidth]{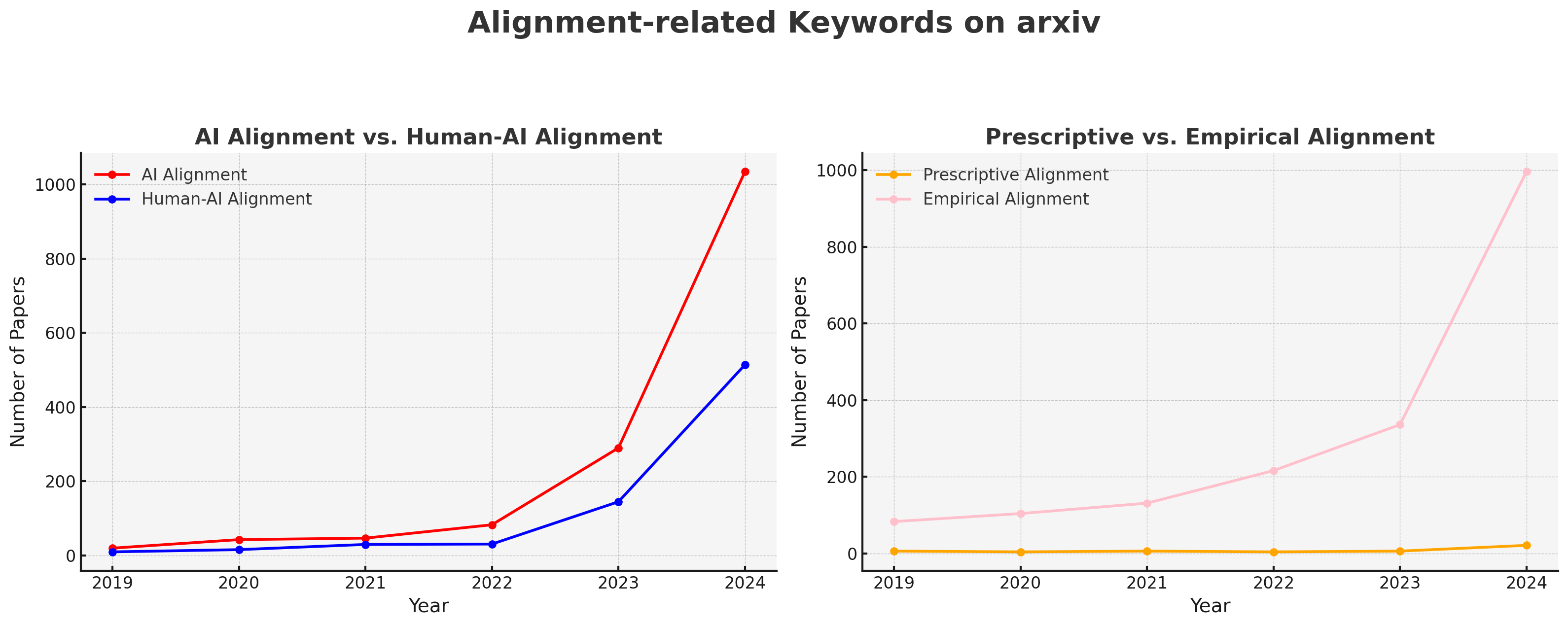}
    \caption{Number of arxiv paper uploads per year with alignment-related keywords (2019-2024). 
    }
    \label{fig:arxiv-trends}
\end{figure*}

\begin{small}
    \begin{quote}
         "If we use, to achieve our purposes, a mechanical agency with whose operation we cannot interfere effectively, [...] we had better be quite sure that the purpose put into the machine is the purpose which we really desire and not merely a colorful imitation of it." \flushright -- \citet{wiener1960automation}, In: \textit{Science}, 131(3410), page~1355.
    \end{quote}
\end{small}
Aligning artificial intelligence (AI) with human goals has shifted from abstract ethics to urgent policy agendas. Regulators see safety and harm prevention as prerequisites for AI's widespread deployment, not optional virtues, see e.g., \citet{chatila2019ieee}.
%
%
Against this background, it comes as no surprise that the challenges of human--AI alignment have sparked a lot of interest in the machine learning (ML) research community. 
In 2024 alone, more than 1500 papers with alignment-related keywords were uploaded to arxiv, see Fig.~\ref{fig:arxiv-trends}. Analyses of github and Hugging Face uploads show similar trends. An alignment benchmark study \cite{kirk2024prism} won a best paper award at the \nth{38} Conference on Neural Information Processing Systems (NeurIPS). 

The position paper track at last year's \nth{41} International Conference on Machine Learning (ICML) alone saw at least four papers \cite{sorensenposition,lindauer2024position,conitzerposition,yangposition} explicitly calling for more, better, or more nuanced human--AI alignment. \citet{lindauer2024position} advocate for a more human-centered approach to automating ML pipelines. Both \citet{sorensenposition,conitzerposition} emphasize the need to align AI systems with diverse and potentially conflicting human interests. 

\citet{yangposition} move beyond a mere human-centered perspective: They recognize the need to adapt AI to environments and self-constraints in \textit{addition} to human intentions. 
For instance, agents have to be aligned with environmental dynamics in order to understand whether the next actions could violate human preferences learned in the first place. Exclusively aligning with human intentions seems too strong of a restriction even for adhering to the very same intentions in the real world.    
%
%
%

Partly inspired by \citet{yangposition}, \textbf{this interdisciplinary position paper argues that empirical alignment's nobility of purpose can disguise statistical flaws}. 

We are convinced that the pursuit of aligning AI with human preferences through observing the latter may inadvertently introduce biases and limitations into these systems. Unlike \citet{yangposition}, we specifically focus on \textit{forward} (a priori) alignment of ML systems in an \textit{empirical} way. That is, we make the case against aligning AI through \textit{observing} human preferences \textit{before} deployment, see definitions~\ref{def:main} through~\ref{def:emp-align} and Section~\ref{sec:forward-emp}. Constructively, we offer three alternatives: Empirical backward (rather than forward) alignment (Section~\ref{sec:backward}) as well as prescriptive (rather than empirical) alignment, both forward (Section~\ref{sec:prescriptive}) and backward (Section~\ref{sec:decoding}), see Table~\ref{tab:fourfold}.

Our stance relies on a statistical perspective, as empirical alignment hinges on a myriad of (implicit and often ignored) statistical assumptions like representative samples, absence of confounders and well-defined populations. We identify several statistical biases as empirical alignment's Achilles heel, see Section~\ref{sec:stats-biases}. Effectively, they skew the alignment goals away from the \say{purpose which we really desire} to a \say{colorful imitation of it} (\citet{wiener1960automation}, page 1355, see above).

Unlike backward empirical alignment, which aligns AI \textit{after} deployment, forward empirical alignment directly leads to these alignment-caused biases being encoded in the AI models. This makes it hard to disentangle them from the originally trained model. Thus, we contend that AI should be -- wherever possible -- explained and altered \textit{a posteriori} rather than empirically aligned \textit{a priori}.
Section~\ref{sec:backward} compares the merits of backward empirical alignment to the perils of forward empirical alignment in greater detail. 
%

Empirical-alignment-induced biases are not only counterproductive for solving novel tasks in non-human-centered environments \cite{yangposition}. More dramatically, these biases restrict an AI system's potential for scientific discovery by confining them to human-centric perspectives. They hinder the understanding of broader and potentially more important phenomena that lie beyond human interests and perception. 

Such anthropocentric biases are not unique to AI alignment. In Section~\ref{sec:anthro-bias}, we draw parallels to the rich literature on anthropocentrism, eventually relating alignment-induced biases to the \textit{observation selection effect} in physics, see Section~\ref{sec:physics}. We learn that AI systems will miss crucial parts of the universe if empirically aligned with those that are perceived as useful for humans or, more generally, that are perceivable by humans at all. This implies that such biases are unresolvable within an observational, i.e., purely empirical framework. In this way, we expose empirical alignment as susceptible to the positivist \say{myth of that which is the
case} \citep[page~vii]{horkheimer_adorno_dda_engl}, credulously relying on observations to accurately represent alignment goals.\footnote{Beyond alignment, such mindless adherence to observations without questioning their origin and relation to the world has led to non-replicability in ML, see the position paper of \citet{herrmann2024position}.} Indeed, following {\citet[page~93-94]{popper:logic}},\say{[t]he empirical basis} does not provide \say{a solid bedrock,} rather,  \say{[\ldots] science rises [...] above a swamp.}
%

These insights motivate our call for prescriptive alignment, which aligns AI systems with transparent axioms rather than human observations. Roughly speaking, our perspective hinges on the insight that \textbf{humans are reasonably good at defining rational axioms, but rather bad at acting according to them}, see \citet{tversky1974judgment,Birnbaum_2016,balagopalan2023judging} and Section~\ref{sec:prescriptive}. 
Recent work by \citet{kirk2024prism} serves as a motivating example: They find that preferences of humans exposed in conversations with large language models (LLMs) depart from preferences pre-stated in a survey.
In Section~\ref{sec:decoding}, we showcase the advantages of prescriptive over empirical alignment in a case study on decoding of large language models.

In summary, \textbf{this position paper cautions against biases arising from forward empirical human--AI alignment. We advocate for \textit{prescriptive} instead of empirical alignment and contend that, where possible, AI should be \textit{a posteriori} explained and modified rather than \textit{a priori} aligned.  
}

\section{Human--AI Alignment: A Taxonomy}

We do not argue against human--AI alignment \textit{per se}, but rather focus on what we perceive as a popular (see Fig.~\ref{fig:arxiv-trends}) yet often unquestioned approach to alignment in the ML community: \textit{forward} and \textit{empirical} human--AI alignment.
We approach this notion by first defining alignment generally.

\begin{definition}[Human--AI Alignment]\label{def:main}
Consider any ML model identified with parameters $ \hat{\theta} \in \Theta$. Denote by $\omega \in \Omega$ an alignment goal. 
Define human--AI alignment as any process involving changes of $\hat{\theta}$ that considers a function of the form  $\Theta \times \Omega \to \mathcal{A}; (\hat{\theta},\omega) \mapsto ||f(\hat{\theta}) - g(\omega)||,$
where $f: \Theta \rightarrow \mathcal{A}$ and $g: \Omega \rightarrow \mathcal{A}$ map to some alignment space $\mathcal{A}$ equipped with some notion of distance $||\cdot||$. 
For instance, this process can be 
    $
     \min_{\hat{\theta} \in \Theta} ||f(\hat{\theta}) - g(\omega)|| 
    $ for fixed $\omega$.\footnote{
Observe this definition's similarity to model misspecification in statistics \cite{white1982maximum}, serving as additional motivation for examining alignment through a statistical prism, see Section~\ref{sec:stats-biases}.}

\end{definition}

We distinguish between \textit{forward} and \textit{backward} human--AI alignment, somewhat reminiscent of \citet[Section 1.2]{ji2023ai}. Forward alignment refers to harmonizing AI systems with human values already during the training phase and \textit{before} deployment. In other words, $\hat{\theta}$ is altered informed by $(\hat{\theta},\omega) \mapsto ||f(\hat{\theta}) - g(\omega)||$ \textit{before} the test phase. 
The approach is more proactive, embedding alignment into learning itself.

Backward human--AI alignment, in contrast, 
changes the parameters $\hat{\theta}$ of a \textit{trained} model 
\textit{during or after} deployment. It includes safety evaluations, governance frameworks, and interpretability methods, see Section~\ref{sec:backward}. 
This approach is rather reactive, focusing on monitoring, adjusting, and managing AI behavior as new risks or misalignments emerge during deployment.

An attentive reader might already have detected the statistical nuance in this distinction: Our temporal cut-off point to distinguish forward from backward is when the model makes inferential statements about the unobserved population for the first time. In other words, we differentiate between AI systems in a purely descriptive state (capable only of making deterministic statements about the observed “training” sample) and in an inferential state (making uncertain statements about the unknown population or unseen “test” sample thereof). 

The motivation for backward alignment in this distinction is straightforward: Before deployment, changes to $\hat \theta$ are driven by training, i.e., by minimizing a loss $\Theta \times \Theta \rightarrow \mathbb R$. After deployment, it is reasonable to assume that the so-found $\hat \theta$ is \say{sufficiently} (Bayes-)optimal w.r.t. this training objective. Subsequent changes to $\hat \theta$ are then motivated by alignment – i.e., adaptation to human alignment goals $\omega$ beyond mere learning of $\theta$. In contrast, forward alignment directly encodes $\omega$ in $\Theta$ while training.  

\begin{table}[t]
\centering
\caption{Fourfold distinction of forward~vs.~backward and empirical vs. prescriptive human--AI alignment and our positions. \\}
\label{tab:fourfold}
\begin{tabular}{l|l|l}
\toprule
\toprule
                     & \textbf{Empirical}                          & \textbf{Prescriptive}          \\
                     & \textbf{Alignment} & \textbf{Alignment} \\
\midrule
\textbf{Forward}     & Contra (Section~\ref{sec:forward-emp})                                           & Pro (Section~\ref{sec:prescriptive})          \\
\midrule
\textbf{Backward}    & Pro (Section~\ref{sec:backward})   & Pro (Section \ref{sec:decoding})   \\
\bottomrule
\end{tabular}
\end{table}

In addition to the \textit{forward-backward} distinction, we define \textit{prescriptive} human--AI alignment and \textit{empirical} alignment as follows, giving rise to our fourfold distinction in Table~\ref{tab:fourfold}.

\begin{definition}[Prescriptive Human--AI Alignment]\label{def:prescr}
    If $\Omega \ni \omega = \tilde \omega(\mathcal{A})$ are set according to \textit{pre-defined axioms} $\tilde \omega$, the process in definition~\ref{def:main} shall be called \textit{prescriptive} human--AI alignment. 
\end{definition}
\begin{definition}[Empirical Human--AI Alignment]\label{def:emp-align}
    If $\Omega \ni \omega = \hat \omega(X)$ are set according to estimators $\hat \omega(X)$ of $\omega$ from \textit{observed human behaviors} $X$, the process in definition~\ref{def:main} shall be called \textit{empirical} human--AI alignment. Here, $X$ denotes a multivariate random variable $X: S \rightarrow \mathbb{R}^q, q \in \mathbb N,$ describing the data obtained from humans, where $S$ is a sample space.  
\end{definition}

This paper's main argumnet hinges on the simple yet far-reaching insight that empirical alignment entails statistical inference: Alignment goals are statistics of observed human behavior, see definition 2.3. The subsequent alignment thus hinges on their statistical properties –  in the words of Wiener (see above) they should resemble “what we really desire and not a colorful imitation of it”. Statistical biases can render $\omega$ such a colorful imitation. 
We will discuss a myriad of such biases in section 3.3 with respect to how these affect the statistical properties of $\hat \omega(X)$.

We emphasize the abstract nature of this fourfold distinction. In practice, the stylized separation between forward and backward alignment is often less strict. Consider, e.g., continual learning \cite{shin2017continual,kirkpatrick2017overcoming,parisi2019continual,lopez2017gradient}. Here, dynamic environments require sequential training and hence also sequential alignment. 

The bulk of recent human--AI alignment research revolves around empirical alignment, see right chart in~Fig.~\ref{fig:arxiv-trends}. 
Examples of \textbf{forward empirical} comprise reinforcement learning from human feedback (RLHF) \cite{christiano2023deepreinforcementlearninghuman,stiennon2020learning, nakano2021webgpt,
ouyang2022training, bai2022training, glaese2022improving} or direct preference optimization (DPO) \cite{rafailov2023dpo,richardson2024logic,tang2024stochastic, afzali2024aligning, wang2024survey, gupta2024context, yin2024sparse, li2024montessori}. 

In \textbf{backward empirical} alignment, post-deployment fine-tuning, interpretability methods, and safety checks allow the system to be continually refined based on new data \cite{sun2019fine_tune_bert,kairouz2021federated_learning}.  

\textbf{Forward prescriptive} alignment has its origins in rule-based AI and expert systems \cite{giarratano1998expert}. It goes beyond mere training on human preferences and instead introduces explicit principles, guidelines, or \say{constitutions} that prescribe what the AI model should or should not do. Note that these pre-defined axioms can harness observational information, too. They can be functions of potential observations. For instance, rule-based rewards by \citet{mu2024rule} increase sample efficiency of RLHF through rule-based alignment. 

Albeit its origins laying in the last century, prescriptive approaches to AI alignment are still popular in state-of-the art models like DeepSeek-R1 \cite{guo2025deepseek}, which was trained on rule-based rewards for reasoning. 
\citet{bai2022constitutional,bai2022training,kyrychenko2025c3ai,findeis2024inverse} propose codifying guidelines for harmlessness and helpfulness via “Constitutional AI”, while \citet{glaese2022improving} integrate targeted human feedback to produce safer dialogue agents. Normative frameworks like Delphi and Social Chemistry 101 collect large knowledge bases of social and moral rules \cite{delphi2021,forbes2020social}. Meanwhile, \citet{hendrycks2020aligning} underscore the importance of embedding shared human values into model alignment protocols.

Lastly, \textbf{backward prescriptive} alignment involves enforcing explicit rule sets 
once the AI system is already deployed, for instance, via
real-time policy-based governance \cite{ge2024trust_ai,bayne2005policy_command,wang2021explainable_ai, miao2021ai_education} or via decoding of language models, see Section~\ref{sec:decoding}.



\section{\textit{Contra} Forward Empirical Alignment}\label{sec:forward-emp}

As this paper cautions against anthropocentric biases arising from \textbf{forward empirical} alignment, we turn to this kind of alignment in more detail. We first emphasize the philosophical underpinnings of our argument in Section~\ref{sec:anthro-bias}, and then -- motivated by the anthropic principle in physics (Section~\ref{sec:physics}) -- discuss concrete statistical biases in Section~\ref{sec:stats-biases}. 

\subsection{The Anthropocentric Bias}\label{sec:anthro-bias}


Anthropocentric thinking, which places humans at the center of the universe, has deep roots in Western thought. In the Judaeo–Christian tradition, the anthropocentric perspective originates from the creation narrative, where humans are depicted as the pinnacle of creation (Genesis 1:27). Historically, this has heavily influenced science. 

However, the anthropocentric interpretation of Genesis 1:27 faced staunch opposition already from within the Jewish tradition itself. One notable critic was Moses Maimonides, a preeminent Torah scholar of the twelfth century AD \cite{lamm1965man,shapiro2003limits}. 
In his seminal work, \textit{The Guide for the Perplexed}, Maimonides emphasized the vastness and complexity of the universe, which, in his view, diminished the centrality of humans. He famously referred to humans as "just a drop in the bucket," evoking Isaiah (40:15), see \citet{dan1989studies}.
Notably, Maimonides' opposition to anthropocentrism was a theological one. He believed that attributing undue importance to humanity was arrogant and a misinterpretation of divine intent. For Maimonides, recognizing the humility of human existence was essential for a proper understanding of God and the natural order. 

It was more than three centuries later when Nicolaus Copernicus, Galileo Galilei, and Isaac Newton established a new \textit{helio}centric model of the universe, literally shifting the center of the universe away from humans. This paradigm shift \cite{kuhn1997structure} paved the way for the scientific revolution in the 16th and 17th centuries. 

Yet, anthropocentric biases still persist in science to this day. The argument is plain and simple: By focusing on explaining phenomena that are useful for humans, we inadvertently miss structures that are not perceived as useful according to current societal values, which are strongly limited in time and generality. In the words of \citet{petersgresham}, "Nature has much more structure than what is useful for humans." Strong positions within evolutionary epistemology even argue we cannot perceive anything that is \textit{not} useful, since our cognitive apparatus is a product of evolution, thus overfitted to useful elements of nature \citep{Lorenz:Mirror,Wuketits:1990}. 

The example of Maimonides shows that even within an overarching anthropocentric dogma, there is room for critical reflection and nuanced positions. 
This motivates our stance on AI alignment. We are well aware of the pressing need to align AI with human safety constraints in the wake of ever more powerful models and in anticipation of Artificial General Intelligence (AGI), see also Section~\ref{sec:alt-view}. However, we assert that empirical alignment is the wrong path towards that goal.
Just like the Torah scholar Maimonides, who firmly believed in the dogma of his times, we do not oppose the paradigm of alignment generally. Quite the contrary, for alignment to work sustainably, we argue, it has to be freed from anthropocentric biases.

\subsection{The Anthropic Principle in Physics}\label{sec:physics}

\begin{figure}
    \centering
    \includegraphics[width=0.9\linewidth, height=4.9cm]{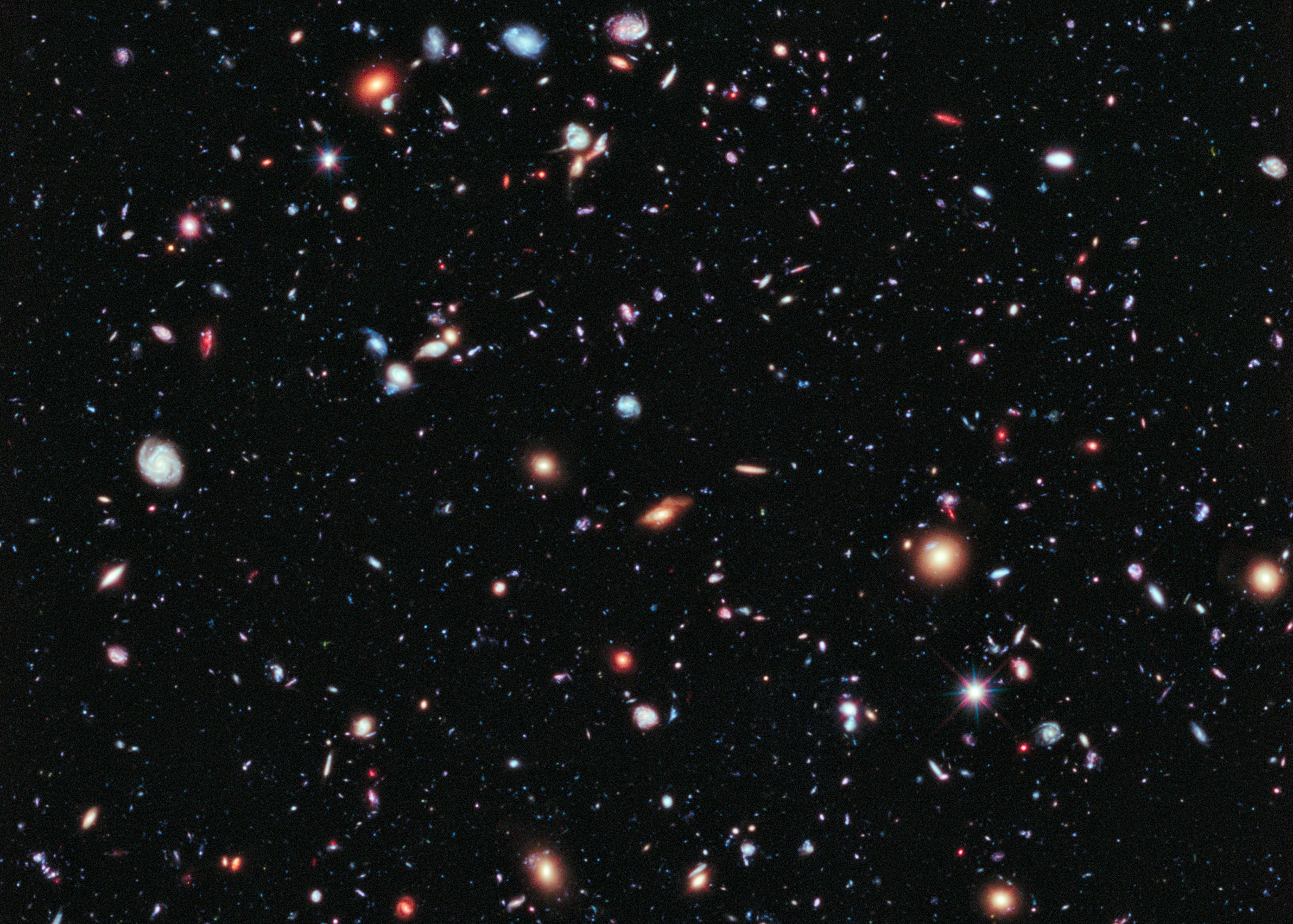}
    \caption{Limits of the human-observable universe: The Hubble eXtreme Deep Field, showing roughly $5500$ galaxies. Source: \url{https://esahubble.org/images/heic1214a/} (acc. 04/15/25)}
    \label{fig:enter-label}
\end{figure}

Modern-day physics is aware of such anthropocentric biases and -- going even further -- the more general anthropic principle, also referred to as "observation selection effect" \cite{bostrom2000observational,bostrom2002anthropic, carter1983principle}. It states that all possible observations of the universe are limited by the fact that they can only be made in a universe capable of developing intelligent life, and is commonly attributed to Robert Dicke \cite{dicke1961dirac,dicke1957gravitation}, building on work by Paul Dirac. For example, constants of nature like the electron charge appear fine-tuned for life because, if they were not, we would not be around to observe them.

Accepting this natural restriction to human reasoning, \citet{bostrom2002anthropic} postulates the self-sampling assumption, which we can directly relate to empirical alignment. It states: \say{One should reason as if one were a random sample from the
set of all observers in one’s reference class} (\citet{bostrom2000observational}, page 57). 

The self-sampling assumption reveals an unresolvable dilemma of empirical alignment. If we define the reference class broad enough to capture all environments relevant to AI deployment, empirical alignment will violate the self-sampling assumption, since the sample is biased towards humans. On the other hand, if the reference class is narrowly restricted to human-related objects, the model can hardly generalize beyond these objectives, see e.g.,\citet{yangposition}.   

Indeed, it was recognized by \citet{schmidhuber2000algorithmic,schmidhuber2002speed} that the anthropic principle provides little insight when the thought experiment is restricted to only one universe. A meaningful theory requires informative priors or alternative universes, see approaches by \citet{schmidhuber2000algorithmic,schmidhuber2002speed,bostrom2002anthropic}. These are all pre-defined axioms, corresponding to a \textit{prescriptive} approach. 

Transferred to empirical human--AI alignment, this insight implies that we have to at least enrich empirical alignment by axiomatic assumptions on the alignment procedure. In Popper's image of empirical science rising \say{above a swamp}, see Section~\ref{sec:intro}, these assumptions are the \say{piles [...] driven down from above into the swamp} 
\citep[page 93-94]{popper:logic}.  
Blindly relying on a sample without additional assumptions on the population where it is drawn from will always lead to biased conclusions about non-human entities. The heart of the matter is that \textbf{making no population-related assumptions implicitly corresponds to making the strongest assumption of all} -- the sample being fully sufficient for the alignment goal. 

The anthropic literature teaches us that we have to explicitly take sample selection probabilities into account, requiring a \textit{statistical} perspective. We thus need to consult the statistical literature on causal and selective inference as well as sampling theory. The following Section does the job.

\subsection{The Statistical Perspective}\label{sec:stats-biases}

The limits of empirical alignment can be best understood from a statistical perspective: Empirical alignment intends to harmonize AI with the intentions of some population of (potential) human agents, constituting $\Omega$ in Definition~\ref{def:main}. This is commonly done by means of a self-selected, thus -- as we learned from the anthropic principle in physics -- distorted sample thereof. This distortion can become manifest in a myriad of statistical biases, 
which particularly have been discussed in the social and survey statistics literature. 

A sole focus on empirical evidence denies the problem of \textit{adequation} \citep{Menges:Adaequation:1982,Grohmann:Method:2000}, which is concerned with the -- \textit{eo ipso} insufficient -- fit between what is, in principle, observable, describable, and analyzable within the framework of our formalization process, and the \say{world-in-itself}. Recall from Definition~\ref{def:emp-align} that empirical alignment relies on estimators $\hat\omega(X)$ from data $X: S \rightarrow \mathbb{R}^q$ obtained from humans, where $S$ is a sample space. The \textit{adequation} problem tells us that we implicitely aim for some latent $X^*: S^* \rightarrow \mathbb{R}^q$, but only have access to its observable counterpart $X$.

\citet{Groves:Lyberg:TSE:2010} concretise some major biases arising from this discrepancy by their \textbf{TSE-(Total Survey Error) concept}. They distinguish between what we call \textbf{population representation biases} and \textbf{structural representation biases}. The first one comprises, e.g., \textbf{biases in the selection frame}. That is, the population from which the sample is taken differs from the population of interest.
That is, $S^* \not = S$. One can distinguish over-coverage ($S^* \subset S$) and under-coverage ($S \subset S^*$). In practice, survey designers often face a mixture of both.

Other instances of the population representation biases are the classical \textbf{sampling error} arising from a random selection of the sample from the underlying population and the \textbf{unit-nonresponses} arising when certain units refuse, or are incapable of, participating in the survey. The latter can distort conclusions because non-respondents' characteristics often differ systematically from those who participate, biasing $S$ away from $S^*$. 

The structural representation biases refer to the content-related part of the analysis and the resulting incomplete reflection of the complex relationships between $X$ and $X^*$. 
This includes 
the \textbf{item-nonresponse bias}, arising when individuals ready to participate in principle refuse to answer certain sensitive questions. \textbf{Response biases} refer to biased data obtained from those that \textit{do} respond, comprising acquiescence bias (a tendency to agree) or primacy and recency effects (a preference for selecting the first or last item in a sequence), see \citet{kaufmann2023challenges}. 
A concrete example of primacy bias in empirical alignment arises in RLHF when presenting more than two options to the human, see e.g. \citet{early2022non}. As was early recognized in the psychology literature, fatigue and distraction can make humans ignore the later (earlier) options and only select from the ones on the top (the bottom) of the user interface \cite{crano1977primacy}. 

A further source of structural representation bias, not explicitly elaborated in the TSE framework by \citet{Groves:Lyberg:TSE:2010}, is the \textbf{omitted variable bias}. It refers to situations when decisive influence factors (like hidden personal characteristics or genetic dispositions) are not accessible to the researcher, for example, for reasons of privacy preservation. This can spuriously enlarge the effect of global variables like sex or age in a merely empirical data-based analysis. 

Note that all the biases listed here are \textit{big data biases} in the sense that they do not vanish with increasing sample size. 
The only exception is the usual statistical uncertainty induced by the sample error.   

 Beyond the TSE, there is a bulk of statistical literature on what \citet{Benjamini2020Selective} calls \say{the silent killer of replicability}: \textbf{Selective inference}. It occurs when hypotheses, models, or features are selected based on the observed data. Such post-selection inference biases arise in numerous scenarios, like variable selection in regression models, multiple hypothesis testing, and adaptive stopping rules in ML \cite{benjamini1995controlling,wasserman2009high,fithian2014optimal,tibshirani2016exact,lee2016exact}. 

Empirical alignment is prone to suffer from selective inference, if the alignment targets $\omega$ are functions of the same observations $X$ being used for finding $\hat{\theta} \in \Theta$ through empirical risk minimization, see definition~\ref{def:main}, as is often the case. If data is sampled anew from a different population for alignment, such trouble can be avoided. In that case, however, the procedure becomes very data hungry. This can pose practical challenges, as human samples are typically burdensome and expensive to acquire \cite{shinn2024reflexion}.
For an illustrative example of selective inference that arises when empirically aligning language models, we refer the interested reader to experimental results in Appendix~\ref{app_sel-inference}.

\textbf{Reflexivity} constitutes an even more fundamental problem in empirical alignment. In addition to the observation selection effect -- which persists in any empirical science, see Section~\ref{sec:physics} --, empirical \textit{human}-AI alignment suffers from a fundamental problem in the \textit{social} sciences: The observant entity (human) coincides with the observing entity (also human). Unlike for celestial bodies (natural sciences), we can expect humans (social sciences) to react to conclusions (in our case, AI alignment) being drawn from observing them, thereby compromising the validity of those conclusions.

This \say{reflexivity problem} \cite{soros2015alchemy} dates back to early work of \citet{morgenstern1928wirtschaftsprognose} and has recently seen some revival in the ML community, recognizing the fact that today's predictions can change tomorrow's population~\cite{perdomo2020performative,hardt2023performative,mendler2020stochastic,miller2021outside}. Examples range from self-negating traffic route predictions to self-fulfilling credits scores.
In empirical alignment, this translates to estimated $\hat \omega$ affecting $X$, from which $\hat \omega$ is learned in the first place. In iterative learning setups, reflexive effects between parameters and data can even occur within the sample \cite{rodemann2025reciprocal,rodemann-bailie-genbounds}.

To make reflexivity more tangible,
consider the popular example of fine-tuning LLMs by RLHF, which implicitly relies on the \say{unrealistic} \cite{carroll2024ai} assumption that human preferences regarding LLM answer quality is static. To say the least, it is certainly plausible that previous LLM answers affect our judgments through, e.g., anchor effects \cite{lieder2018anchoring}. Very recently, \citet{carroll2024ai} took steps to address such feedback loops by drafting AI alignment as a dynamic Markov Decision Process (MDP), taking into account that our preferences can change by interacting with changing AI systems. \citet{mitelut2023intent} go further and argue entirely against alignment to human intent, since AI systems can reshape the latter, see also \citet{gabriel2020artificial}.

\textbf{Causal misrepresentation} refers to misleading associations that do not reflect causal connection between variables. Such correlations occur in observational data, on which empirical alignment mostly still hinges on -- with the notable exception of experimental methods in the assurance literature, see \citet[§4.3]{ji2023ai}. 

In our setup, let $X = (Y,Z)$ with $Y: S_Y \rightarrow  \mathbb{R}^q$ and $Z: S_Z \rightarrow \mathbb{R}^q$ random variables denoting such experimental data. Further, assume our alignment goal $\omega$ consists of the causal effect of $Z$ on $Y$, e.g., answer length (Z) on answer quality rated by humans (Y) for LLM replies. We do not know the actual (true) causal effect $\omega$, but we observe human ratings of the quality Y and LLM answer lengths Z and can compute an estimator $\hat \omega(Y,Z)$, e.g. the Bravais-Pearson correlation coefficient: 
\[
\hat \rho(Y,Z)
\;=\;
\frac{\displaystyle \operatorname{Cov}(Y,Z)}{\displaystyle \operatorname{Var}(Y)\, \operatorname{Var}(Z)}.
\]
Causal misrepresentation bias implies there is an additional variable $C$, e.g. educational background, that affects both Z and Y. For instance, people with college degrees might tend to prefer longer responses (effect of $C$ on $Y$). At the same time, college-educated prompters might ask more complicated questions triggering longer replies (effect of $C$ on $Z$). This implies a spurious correlation between $Y$ and $Z$. Our estimator $\hat \omega$ is thus biased away from the true $\omega$. It falsely captures the relation between answer length and quality, while the actual cause of both (educational background) is not taken into account.

Statistical science offers sound remedies for many of the above mentioned biases, see \citet{nalenz2024learning,rodemannnot} for instance. They typically involve additional assumptions about the population like \say{absence of unobserved confounders.} Essentially, these assumptions are \textit{prescriptive} elements, which we advocate for. Even if they are self-evident as in the case of confounders, we strongly encourage making them explicit. 

This increases transparency not only for humans but also for AI, thus addressing potential unintended misalignments due to miscommunication. \textbf{If AI systems fail to understand that the actual goal is population-alignment rather than sample-alignment, consequences can be dire}, see \citet[pages 122-125]{bostrom2014superintelligence} and \citet[Chapter 8]{harari2024nexus}. For instance, spurious correlation between race and occupation status in alignment samples might lead to discrimination through aligned AI Systems.  

While these statistical remedies address known epistemic uncertainties \cite{hullermeier2021aleatoric} (\say{known unknowns}) about the population, there are potentially many more unaddressed sources of uncertainty (\say{unknown unknowns}), see e.g., \citet{rodemann2021accounting,rodemann2022accounting,rodemann2024imprecise}. This especially holds for statistical representation of \textit{preference} data, essential to measuring human intentions, as we detail in the next Section.

\section{\textit{Pro} Forward Prescriptive Alignment}\label{sec:prescriptive}

The theory of preference(s) (relations) originates in modeling rational agents within the foundations of decision theory. It finds ever broader areas of application in modern ML (e.g.,~\citet{pl2003,JMLR:v24:22-0902,pmlr-v216-jansen23a,JMLR:v22:18-546,jansen2024statistical,rodemann2023approximately,rodemann2023all,rodemann2025reciprocal,dietrich2024semi,kim2024queueing}). In particular, the subfields of \textit{preference elicitation} (e.g.,~\citet{haddaway,baarslag2015,mukherjee2024optimal}) and \textit{preference aggregation} (e.g., \citet{ehl2012,JANSEN201849,pmlr-v235-zhang24u}) play decisive roles here. 
In what follows, we support our case \textit{pro} forward prescriptive alignment by findings from these two subfields. 

 The term \textit{preference elicitation} refers to the systematic retrieval of initially unknown (human) preferences, often achieved by successively (and adaptively) presenting queries to the agent under consideration. It is an excellent illustration of the superiority of a prescriptive understanding of alignment (using \textit{pre-defined axioms} $\tilde \omega$ as alignment targets $\omega \in \Omega$, see Definition~\ref{def:prescr}) over a purely empirical understanding of alignment (Definition~\ref{def:emp-align}): It is easy to find \textit{pre-defined axioms} on the nature of preference structures modeling rational behavior that meet broad social consensus. 
 -- and it is precisely these supposedly rational axioms that are often violated in empirical studies. 
 
 These violations, however, are often due to lacking oversight rather than being conscious and intended choices of the agent. At the same time, integrating rationality axioms in elicitation strategies is an instance of forward prescriptive alignment: The axioms are agreed upon \textit{before} the model is deployed.  

Probably the most prominent example of a rationality axiom in the context of preference elicitation is that of \textit{transitivity}: If an agent prefers $a$ over $b$ and $b$ over $c$, then the preference of $a$ over $c$ should follow. While there is strong theoretical support of transitivity via \textit{ money-pump arguments} (e.g., \citet{ALDRED01012003}), numerous studies (e.g.,~\citet{birnbaum1999evidence,Birnbaum_2016,birnbaum2008experimental,segal2021,guadelup2020}) show that agents often behave intransitively or even cyclic, although they fully agree with transitivity as a rationality axiom. 

This happens in particular when the choice situation is dynamic and a large number of different options for action are available. The intransitivity of the agent's behavior in such cases is rather an expression of their limited comprehension than an ideal of action to be elevated to a principle. Consequently, the naive demand for a purely empirical alignment here would translate to an AI that only imitates the agent's lack of oversight instead of pursuing and supporting their actual interests. 

A simple \textit{prescriptive} alignment strategy that prevents the occurrence of intransitive preference patterns by construction was proposed by \citet{jbas2022}: Instead of asking the agent about all possible pairwise comparisons step by step, the \textit{transitive hull} of the elicited preferences is formed after each query. The next pair to be presented is then selected from the remaining pairs, \textit{excluding} those pairs that would follow anyway due to transitivity, ensuring the final result to be transitive. 

Beyond elicitation procedures, another problem of purely relying on empirical alignment arises in the context of \textit{preference aggregation}. This term refers to the problem of defining procedures that aggregate arbitrary \textit{profiles} (fixed-size tuples) of preference relations to one collective preference, in a way that is conceived fair. For instance, preference aggregation plays a crucial role in benchmarking problems under multiple evaluations (e.g., \citet{ehl2012,mptbw2015,iclr-2024,pmlr-v235-zhang24u}), where one aims to aggregate several rankings of ML algorithms, e.g., obtained by domain experts.

Infamously, the attempt to define (full-domain) aggregation procedures for preference profiles is generally doomed to failure already under very weak axiomatic conditions. This follows from Arrow's impossibility theorem \citep{arrow1950}, which was 
discussed in \citet{conitzerposition} as a major challenge in aligning AI with diverse human feedback. 

A common way to get around this impossibility is to retreat to the Pareto front of the aggregation problem, i.e. to base analyses on the (undominated elements of the) \textit{partial} preference order arising as the intersection of the preference orders in the respective profile. 
However, this strategy, too, has serious disadvantages: Retreating to the Pareto front -- given some of the preference orders in the profiles originate from some (latent) real-valued score -- results in significant information loss and thus inefficiency, see. e.g., \citet{mdai, farrow2009almost}. 

Both challenges just sketched -- aggregation and Pareto-analysis -- essentially originate from an implicit assumption of purely empirical forward alignment. 

Accordingly, both challenges can potentially be avoided by a prescriptive approach: In the context of aggregation, this would imply that the aggregate does not have to correspond directly to a function of the preference orders of the individual agents. Instead, external factors could also be taken into account in the aggregation, allowing the rigid framework of the impossibility theorem to be left behind. In the light of the second problem, the inefficiency of the Pareto front can also be remedied: As demonstrated in e.g., \citet{kreuter2013improving,jbas2022}, in the process of eliciting preferences, paradata about preference intensity (i.e., latent cardinal information) can be implicitly collected, allowing the full available information to be included. Under a prescriptive alignment paradigm, this data -- not directly related to targeted human behavior -- could be used to obtain significantly more information-efficient improvements on the Pareto front.

\begin{table*}[t]
\centering
\small
\caption{Empirical vs. prescriptive alignment of language models: Decoding strategies Contrastive Search (CS) and DoubleExp are compared across three datasets using two alignment metrics: QText (\textit{prescriptive}) and MAUVE (\textit{empirical}) for \textbf{Automatic Evaluation}. \textbf{Human Evaluation} results indicate the percentage of evaluators favoring each strategy based on perceived semantic coherence and fluency. Human selections align with QText (\textit{prescriptive}), see \citet{garces-arias-etal-2024-adaptive}. All results are reported as percentages.}

\label{tab:autom_eval_main}
\begin{tabular}{|l|cc|cc||cc|cc|}
\hline
& \multicolumn{4}{c||}{\textbf{Automatic Evaluation}} 
& \multicolumn{4}{c|}{\textbf{Human Evaluation}} \\
\cline{2-9}
\multirow{3}{*}{\textbf{Dataset}} 
& \multicolumn{2}{c|}{QText (\textit{prescriptive}) ↑} & \multicolumn{2}{c||}{MAUVE (\textit{empirical}) ↑} 
& \multicolumn{2}{c|}{Semantic coherence ↑} & \multicolumn{2}{c|}{Fluency ↑} \\
& CS & DoubleExp & CS & DoubleExp & CS & DoubleExp & CS & DoubleExp \\
\hline
Wikinews    & 91.95 & 82.79 & 84.14 & 90.65 & 73.00 & 27.00 & 61.00 & 39.00 \\
Wikitext    & 87.64 & 81.50 & 77.97 & 84.07 & 57.00 & 43.00 & 60.50 & 39.50 \\
BookCorpus  & 88.71 & 82.12 & 84.74 & 85.66 & 64.50 & 35.50 & 61.00 & 39.00 \\
\hline
\hline
All         & 89.68 & 82.22 & 82.82 & 87.16 & 66.00 & 34.00 & 59.00 & 41.00 \\
\hline
\end{tabular}
\vspace{1.5cm}
\end{table*}

\section{\textit{Pro} Backward Empirical Alignment}\label{sec:backward}
Another argument versus forward empirical alignment stems from the pursuit of transparency.
While most ML models are powerful predictors their decision making process typically is not human intelligible -- with the exception of intrinsically interpretable models. Transparency, however, is desired to understand model decisions especially in high-stakes environments -- let alone legal requirements like the EU AI Act. 

We argue that forward empirical alignment adds to the opacity of models. In RLHF, for example, the reward function is typically approximated by a deep neural network trained on observed human behavior \cite{christiano2023deepreinforcementlearninghuman}, and hence the (often nontransparent) policy is optimized using feedback from a nontransparent reward model. More importantly, though, any forward empirical alignment encodes biases into the model that cannot be disentangled from dependencies in the training data as discussed in Section~\ref{sec:stats-biases}, which further impedes model understanding. 

Thus, we argue for more transparent alignment. Prescriptive alignment techniques check this box by explicitly stating the axioms a model shall be aligned to. If the latter is impractical, however, we offer changes of an ML model's parameters $\hat{\theta} \in \Theta$ informed by interpretable ML (IML) as potential strategy. We call this \textit{backward} alignment.  Modifications to the model architecture, by contrast, are another valid option but per definition not part of the alignment process of the original model.

There is a vast literature on IML methods for classical ML (e.g., \citet{molnar2022,ribeiro2016LIME, SHAP_LundbergL17, greenwell2018pdp, covert20_sage}) as well as those dedicated to explain decision making in RL (e.g., \citet{XRL_survey_Piutta, XRL_survey_milani,madumal2023_causallens, topin_2019_apg, Olson_2021_counterfactual, RL_critical_states}).

Such methods typically provide explanations of ML models during or after deployment that are subsequently inspected by human observers. Upon inspection, model explanations can \textit{a posteriori} \textit{assure} that the model works as intended. On the contrary, the observer can infer that the model behaves against their intentions and imply changes on the parameters $\hat{\theta} \in \Theta$ without ever explicitly stating their preferences and while never providing universally applicable axioms. Thus, we merely observe human preferences implicitly and call the latter scenario \textit{empirical backward alignment}. 

Note that -- while inherently prescriptive -- the choice of the (I)ML method is not yet part of the alignment process. As we argue for transparent alignment, we contend that such \textit{a posteriori} informed backward alignment achieves this due to the more explicit nature of the performed changes unlike forward empirical alignment. Moreover, the strategy comprises a rather static choice situation which can mitigate the issue of irrational behavior in empirical alignment raised in Section~\ref{sec:prescriptive}.

As a tangible example of backward empirical alignment, assume a model was trained to classify images into “husky” and “wolf”. While highly predictive, in the example from \citet{ribeiro2016LIME}, the model explanation shows white pixels (i.e., snow) to be an influential feature. Hence, you find the model to be unaligned with prior knowledge, as snow is not the crucial difference between both classes, and you would not trust the model to generalize well. Based on this insight, you may want to change $\hat{\theta}$ by retraining the model on more diverse training data. Note that any subsequent model change will always be informed by model behavior after the first deployment, and hence — once this knowledge was introduced — the process shall be considered backward.











\section{\textit{Pro} Backward Prescriptive Alignment}\label{sec:decoding}

An illustrative example of backward alignment in language models emerges in the choice of decoding strategies for autoregressive text generation. These strategies specify how each subsequent token is selected from the model’s probability distribution over tokens at every inference step, thereby exerting a critical influence on the quality of the generated text.

Various automatic metrics have been introduced to evaluate text generated by different decoding strategies. Among them are several \textit{prescriptive} ones like coherence, diversity or QText (harmonic mean of these latter), and \textit{empirical} ones such as MAUVE \cite{pillutla2021mauvemeasuringgapneural}. For a detailed overview and technical description, we refer to Appendix~\ref{app:prescriptive-metrics} and~\ref{app:empirical-metrics}. While coherence assesses how likely the generated text is given the prompt, diversity measures lexical repetition rates \cite{su2022empiricalstudycontrastivesearch}.  
In contrast, MAUVE measures how closely machine-generated text aligns with empirical samples of human-written text by comparing their distributional “fingerprints” in a latent representation space. Technically, it calculates the Kullback–Leibler divergence between the two distributions. Higher MAUVE scores correspond to lower divergence, indicating greater similarity to human-produced text.

Choosing a decoding strategy to align the model output with human-generated text is an obvious example of \textit{backward} alignment: Decoding strategies are chosen \textit{after} the language model has been trained, allowing for fine-tuning towards human preferences. Relying exclusively on MAUVE to select or evaluate decoding strategies constitutes \textit{backward empirical alignment}: the quality of a strategy is evaluated based on how well its text generations align with observed human-written text samples. This methodology has notable limitations, as our illustrative case study reveals: Based on experimental results in~\citet{garces-arias-etal-2024-adaptive,garces-arias-etal-2025-decoding}, we compare the quality (assessed by human evaluators) of text generated by the same model (GPT2-XL \cite{radford2019language}) using two different decoding strategies, namely CS (contrastive search, \citet{su2022empiricalstudycontrastivesearch,su2023contrastivesearchneedneural}) and DoubleExp \cite{garces-arias-etal-2024-adaptive}. In one scenario, the decoding strategy is empirically aligned via MAUVE. In the other scenario, we use QText, a prescriptive metric, to choose the decoding method. 

Table~\ref{tab:autom_eval_main} highlights discrepancies between high MAUVE scores and human judgments of final text quality. In other words, if we choose a decoding strategy \textit{empirically} through MAUVE, the output is judged to be of lower quality than the one generated by a \textit{prescriptively} chosen decoding strategy. While DoubleExp consistently achieves higher MAUVE scores than CS, it is rejected by human evaluators, which prefer CS for its perceived semantic coherence and fluency. 

In a nutshell, these findings suggest that humans favor text from a model that was not aligned to empirically observed human-generated text but to a prescriptive metric. For further results, we refer to Tables \ref{tab:autom_eval} and \ref{tab:human_eval} in Appendix~\ref{app:further-results}. 
%
Conclusively, we emphasize that these findings are not limited to specific decoding methods like CS, see comprehensive studies in \citet{su2022empiricalstudycontrastivesearch,garces-arias-etal-2024-adaptive,arias2024towards,garces-arias-etal-2025-decoding,GUARD-25}. 

\section{Alternative Views}\label{sec:alt-view}

In this Section, we present three alternative views -- \textit{pro} forward empirical alignment, \textit{contra} prescriptive alignment, and \textit{contra} alignment altogether.

\subsection{Pro Forward Empirical Alignment} 
While statistical caution is appropriate, there might be other -- potentially superior -- reasons to \textit{do} align AI with human preferences in a forward empirical way. Often, ML's main and only goal is good prediction. Strong cases for empirical alignment are especially concerned with such "performance" goals (see \citet{reward_learning_from_human_pref_borja}).

A prime example is \textit{InstructGPT} \cite{ouyang2022instructgpt}, an LLM trained with a "technique [that] can align to a specific human reference group for a specific application" (p. 18). The authors name improved performance and cost-effectiveness over larger models as key benefits of their approach. \citet{stiennon2020learning} report similar benefits of smaller, empirically forward aligned models for the clearly defined task of text summarization. Potentially reduced model size while maintaining predictive performance is further desirable for the widespread deployment of ML models on less powerful hardware. We refer to Appendix~\ref{app:alt-view} for a similar argument in defense of forward empirical alignment from a benchmarking perspective.

\subsection{Contra Prescriptive Alignment}

In this paper, we argue that prescriptive alignment should be favored over empirical alignment, since the latter suffers from serious statistical shortcomings; nevertheless, prescriptive alignment is not without its own limitations. How to agree on predefined axioms $\omega$ that govern prescriptive alignment, is highly relevant and non-trivial. People often disagree with each other on which principles are normatively desirable. Prescriptive alignment is only valuable if one accepts the principles that have been agreed upon in the first place.

Nevertheless, there are two clear advantages over a naive empirical alignment process. Firstly, alignment becomes more transparent and thus more objective through a prescriptive approach: If you agree with the pre-defined normative principles, you can check whether the AI acts in accordance with these principles and, if in doubt, intervene. This seems hardly possible for empirical alignment, where we cannot distinguish whether the AI breaks the rules or the human is just inconsistently violating principles they would accept in theory. Secondly, prescriptive alignment offers the possibility of AI systems acting according to normative principles controlled by humans, which humans themselves may not be able to implement in their concrete behavior.

\subsection{A Case Against Alignment Altogether} With the very same principled arguments from Section~\ref{sec:forward-emp}, one might as well arrive at the reasonable position of giving up alignment entirely. 
Our offered alternatives (backward and/or prescriptive alignment) increase transparency and decrease negative effects of statistical biases.
Yet, they cannot resolve the fundamental dilemma of the observation selection effect: As detailed in Section~\ref{sec:anthro-bias}, any alignment to humans will necessarily bias the AI away from other entities in nature. Fundamentally opposing alignment might thus be particularly justified if the AI's goal is scientific inference, e.g., by means of IML.

While we considered explicit changes to a model informed by IML as \textit{backward alignment}, the use of IML can end at the stage of model explanation as a tool for scientific inference \cite{Freiesleben_2024, ewald2024_fi, molnar_iml_dgp_2023, könig2024dip} -- typically about the data generating distribution. From a statistician's perspective, we deem such inference as valuable in itself and further argue that principled explanations aid in understanding model decisions. \citet{covert20_sage} introduce SAGE values for so called \textit{global feature importance} that -- under certain conditions -- represent mutual information between inputs and output or conditional output variance, \citet{könig2024dip} extend similar insights to feature dependencies and \citet{Freiesleben_2024} define a general framework to design and use IML methods for scientific inference grounded in statistical learning theory.

We contend that a bias from any form of alignment encoded in the ML model can confound the relations between the model variables and as a result diminish the potential for scientific discovery.
%
This argument is further underlined by findings from RL. 
\citet{alphagozero_silver} show that the famous \textit{AlphaGo} model trained from observations of human play is inferior to \textit{AlphaGo Zero} trained \textit{without} human knowledge. Moreover, \citet{schut2023bridginghumanaiknowledgegap} use \textit{concept-based} explanations to extract sequences of actions in chess from an AI trained without human oversight that go beyond human skill level. 

It thus can be argued that not only forward empirical, but any kind of alignment -- in the language of \citet{schut2023bridginghumanaiknowledgegap} -- biases the \textit{machine representational space} towards the \textit{human representational space}. It introduces or exacerbates an anthropocentric bias that can substantially reduce AI's ability to learn concepts beyond human knowledge.





\section{Conclusion}


AI alignment is a double-edged sword. Done well, it makes AI safer. Done poorly, it biases models and limits their potential for discovery.
This paper developed a nuanced statistical perspective on this trade-off. It cautioned against forward empirical alignment unless strictly necessary. 

We emphasized the statistical caveats of common empirical alignment practices (Section~\ref{sec:forward-emp}) and scrutinized (in)consistencies of observed human preferences (Section~\ref{sec:prescriptive}). We showed that \textit{forward} empirical alignment is especially prone to \say{locking in} statistical biases during training. What is more, we constructively discussed alternatives like \textit{prescriptive} (Section~\ref{sec:prescriptive} and~\ref{sec:decoding}) and \textit{backward} (Section~\ref{sec:backward} and~\ref{sec:decoding}) alignment. We further provided practical guidance by concrete examples like decoding of language models (Section~\ref{sec:decoding}).

Along the way, we discovered a fourfold taxonomy -- distinguishing forward vs. backward, empirical vs. prescriptive (Table~\ref{tab:fourfold}) -- of alignment methods that might be of independent interest. Our hope is that this taxonomy and the accompanying arguments guide the ML community towards (more) statistically informed alignment. 
We wish to stimulate a constructive debate on the trade-offs between these different approaches to alignment, see also the contrasting opinions in Section~\ref{sec:alt-view}.  

To invoke Norbert Wiener, whose famous quote on the alignment problem (see Section~\ref{sec:intro}) motivated this whole paper, once again:

\begin{small}
    \begin{quote}
        \say{Moreover, if we move in the direction of making machines which learn and whose behavior is modified by experience, we must face the fact that every degree of independence we give the machine is a degree of possible defiance of our wishes.} \flushright -- Norbert Wiener, 1949, \citet[acc. 01/15/25]{wiener-nyt}
    \end{quote}
\end{small}

As AI grows more autonomous, \say{who aligns whom?} will become a central ethical question. Our position is clear: Alignment should be governed by principles, not just by unreflected mirroring of observations.


\section{Statement on Societal Impact}

This paper warns against the uncritical and naive use of forward empirical alignment, arguing that such alignment introduces statistical biases and anthropocentric constraints. We propose alternatives: prescriptive alignment and backward adjustments. These latter ensure transparency and prevent AI from merely imitating (potentially irrational and inconsistent) human behavior at the cost of better reasoning. Misalignment can lead to significant societal risks, as touched upon in Section~\ref{sec:stats-biases}. 

Future AI policy should thus consider the biases discussed in this position paper. Regulation should not blindly enforce empirical alignment, but instead, demand transparency in how models are aligned. This means documenting alignment assumptions, ensuring explainability, and regularly auditing statistical biases in training data.

The ethical stakes are high. If AI is aligned to flawed human preferences, it may amplify societal biases rather than correct them. 

At worst, it could reinforce harmful power structures, see the example of racial biases in Section~\ref{sec:stats-biases}. Conversely, dismissing empirical alignment entirely might create models that fail to serve human needs. We advocate a middle ground: rigorous, principle-based alignment that minimizes bias while allowing AI to generalize beyond human-imposed limits. 

Particularly, by pushing for a middle ground, we object to the misinterpretation of this paper's arguments as a call against alignment altogether. We discussed this as an alternative view in Section~\ref{sec:alt-view}. Abandoning alignment entirely poses the severe and existential risk of loosing human control over AI systems, potentially giving rise to several \say{doom} scenarios, see also Section~\ref{sec:anthro-bias}.







\newpage
\appendix
\onecolumn

\section{Metrics for Evaluating Decoding of Language Models}

In what follows, we discuss the prescriptive and empirical metrics used for language model alignment via decoding strategy selection in Section~\ref{sec:decoding}.

\subsection{Prescriptive Metrics}\label{app:prescriptive-metrics}

\paragraph{Coherence.}

Proposed by \citet{su2022empiricalstudycontrastivesearch}, the coherence metric is defined as the averaged log-likelihood of the generated text conditioned on the prefix text as

$$
\operatorname{Coherence}(\hat{\boldsymbol{x}}, \boldsymbol{x})=\frac{1}{|\hat{\boldsymbol{x}}|} \sum_{i=1}^{|\hat{\boldsymbol{x}}|} \log p_{\mathcal{M}}\left(\hat{\boldsymbol{x}}_i \mid\left[\boldsymbol{x}: \hat{\boldsymbol{x}}_{<i}\right]\right)
$$

where $\boldsymbol{x}$ and $\hat{\boldsymbol{x}}$ are the prefix text and the generated text, respectively; [:] is the concatenation operation and $\mathcal{M}$ is an external language model, namely OPT (2.7B) \cite{zhang2022optopenpretrainedtransformer}.

\paragraph{Diversity.}

Proposed by \citet{su2022empiricalstudycontrastivesearch}, the diversity metric aggregates $\mathrm{n}$-gram repetition rates: $$\text{DIV}=\prod_{n=2}^4 \frac{\mid \text { unique } \mathrm{n} \text {-grams }\left(\mathrm{x}_{\text {cont }}\right) \mid}{\mid\text { total } \mathrm{n} \text {-grams }\left(\mathrm{x}_{\text {cont }}\right) \mid}$$ A low diversity score suggests the model suffers from repetition, and a high diversity score means the model-generated text is lexically diverse.

\paragraph{QText.}

QText \cite{garces-arias-etal-2025-decoding} is given by the harmonic mean of rescaled coherence and diversity:

$$
\text{QText} = \frac{2}{\frac{1}{\text{COH}} + \frac{1}{\text{DIV}}}*100,
$$

where 

\[
\operatorname{COH} = \frac{\text{Coherence} - \min(\text{Coherence}) + 1}{\max(\text{Coherence}) - \min(\text{Coherence}) + 1}.
\]

QText values close to 100 indicate high-quality text generation, while values approaching zero reflect low-quality outcomes.

\subsection{Empirical Metrics}\label{app:empirical-metrics}

\paragraph{MAUVE.} MAUVE \cite{pillutla2021mauvemeasuringgapneural} is a metric designed to quantify how closely a model distribution \(Q\) matches a target distribution \(P\) of human texts. Two main types of error contribute to any discrepancy between \(Q\) and \(P\): 
\begin{itemize}
    \item \textbf{Type I Error:} \(Q\) assigns high probability to text that is unlikely under \(P\).
    \item \textbf{Type II Error:} \(Q\) fails to generate text that is plausible under \(P\).
\end{itemize}

\noindent
These errors can be formalized using the Kullback--Leibler (KL) divergences \(\mathrm{KL}(Q \!\parallel\! P)\) and \(\mathrm{KL}(P \!\parallel\! Q)\). If \(P\) and \(Q\) do not share the same support, at least one of these KL divergences will be infinite. To address this issue, \citet{pillutla2021mauvemeasuringgapneural} propose measuring errors through a mixture distribution
\[
R_{\lambda} \;=\; \lambda P \;+\; (1-\lambda)\,Q
\quad\text{with}\;\lambda \in (0,1).
\]
This leads to redefined Type I and Type II errors given by
\[
\mathrm{KL}\!\bigl(Q \!\parallel\! R_{\lambda}\bigr)
\quad \text{and} \quad
\mathrm{KL}\!\bigl(P \!\parallel\! R_{\lambda}\bigr),
\]
respectively.

\bigskip
\noindent
By varying \(\lambda\) and computing these two errors, one obtains a \emph{divergence curve}
\[
\mathcal{C}(P, Q)
\;=\;
\left\{
\left(
\exp\!\bigl(-c\,\mathrm{KL}(Q \!\parallel\! R_{\lambda})\bigr),\;
\exp\!\bigl(-c\,\mathrm{KL}(P \!\parallel\! R_{\lambda})\bigr)
\right)
\,:\,
R_\lambda = \lambda P + (1-\lambda) Q,\;
\lambda \in (0,1)
\right\},
\]
where \(c > 0\) is a hyperparameter that controls the scaling.

\bigskip
\noindent
Finally, \(\operatorname{MAUVE}(P, Q)\) is defined as the area under the divergence curve \(\mathcal{C}(P, Q)\). Its value lies between 0 and 100, with higher values indicating that \(Q\) is more similar to \(P\).

\section{Further Results on Decoding Alignment of Language Models}\label{app:further-results}

\begin{table*}[t]
\centering
\small 
\caption{Automatic evaluation results: Comparison of Contrastive Search (CS) and DoubleExp across Wikinews, Wikitext, and BookCorpus using two automatic metrics: QText and MAUVE. DoubleExp outperforms CS in terms of MAUVE, while CS outperforms DoubleExp in terms of QText across all three datasets.}
\label{tab:autom_eval}
\begin{tabular}{|l|cc|cc|}
\hline
\multirow{2}{*}{Dataset} & \multicolumn{2}{c|}{QText (\textit{prescriptive}) ↑} & \multicolumn{2}{c|}{MAUVE (\textit{empirical}) ↑} \\

 & CS & DoubleExp  & CS & DoubleExp \\
\hline
Wikinews  & 91.95 & 82.79 & 84.14 & 90.65 \\
Wikitext  & 87.64 & 81.50   & 77.97 & 84.07  \\
BookCorpus     & 88.71 & 82.12  & 84.74 & 85.66 \\
\hline
\hline
All       & 89.68 & 82.22 & 82.82 & 87.16  \\
\hline
\end{tabular}
\end{table*}

\begin{table*}[t]
\centering
\caption{Human evaluation results: Human ratings of Contrastive Search (CS) and DoubleExp across Wikinews, Wikitext, and BookCorpus, focusing on perceived semantic coherence and fluency of the generated text. DoubleExp is consistently rejected by human evaluators.}
\label{tab:human_eval}

\resizebox{\textwidth}{!}{%
\begin{tabular}{|l|c|c|c|c|c|c|}
\hline
\multirow{2}{*}{Dataset} & \multicolumn{3}{c|}{Semantic coherence} & \multicolumn{3}{c|}{Fluency} \\
 & CS is better & CS and DoubleExp are similar & DoubleExp is better & CS is better & CS and DoubleExp are similar & DoubleExp is better \\
\hline
Wikinews & 56\% & 34\% & 10\% & 32\% & 58\% & 10\% \\
Wikitext & 34\% & 46\% & 20\% & 29\% & 63\% & 8\% \\
BookCorpus & 49\% & 31\% & 20\% & 32\% & 58\% & 10\% \\
\hline
\hline
All & 48\% & 36\% & 16\% & 28\% & 62\% & 10\% \\
\hline
\end{tabular}
}

\end{table*}

We provide additional details on the experimental results presented in Section~\ref{sec:decoding}, see also \citet{garces-arias-etal-2024-adaptive,garces-arias-etal-2025-decoding}.
Tables~\ref{tab:autom_eval} and~\ref{tab:human_eval} provide a more granular breakdown of the empirical versus prescriptive alignment methods evaluated in Table~\ref{tab:autom_eval_main} in the main paper. Table~\ref{tab:autom_eval} presents quantitative comparisons across various datasets and evaluation metrics, highlighting the alignment discrepancies when relying on empirical metrics like MAUVE versus prescriptive metrics such as QText. Table~\ref{tab:human_eval} provides a more detailed breakdown of the human evaluation results reported in Table~\ref{tab:autom_eval_main} in the main paper. The latter summarized the human preferences by reporting the share of text generations that was preferred, while Table~\ref{tab:human_eval} additionally shows indifference votes. 
The results reinforce the finding that empirical alignment methods may not necessarily correspond to human-perceived text quality. This further substantiates our argument for a more principle-driven approach to AI alignment. 
To reflect the share of human raters favoring each decoding strategy, as depicted in Table \ref{tab:autom_eval_main} in the main paper, we apply the following scoring approach:

$$
\text { Score }_{\mathrm{CS}}=\frac{\#(\mathrm{CS} \text { is better })+0.5 \times \#(\text { tie })}{\#(\mathrm{CS} \text { is better })+\#(\text { tie })+\#(\text { DoubleExp is better })} 
$$

$$
\text { Score }_{\mathrm{DoubleExp}}= 100 - \text { Score }_{\mathrm{CS}}.
$$

\section{Selective Inference in Decoding Alignment of Language Models} \label{app_sel-inference}

\begin{table*}[ht]
\centering
\resizebox{0.7\textwidth}{!}{
\begin{tabular}{lccccccc}
\toprule
Dataset & Truncation & \multicolumn{2}{c}{\# Examples} & \multicolumn{3}{c}{MAUVE(\%)↑} & Preferred  \\
\cmidrule(lr){3-4} \cmidrule(lr){5-7}
 & length & CS & ACS & CS & ACS &  $\Delta$ & Method \\
\midrule
Wikinews & 64 & 1939 & 2000 & 87.42 & 85.79 & -1.63 & CS \\
 & 96 & 1920 & 2000 & 81.11 & 88.13 & 7.02 & ACS \\
 & 128 & 1859 & 1977 & 84.14 & 85.39 & 1.25 & ACS \\
 & 160 & 1684 & 1824 & 84.86 & 85.78 & 0.92 & ACS \\
 & 192 & 1447 & 1617 & 85.23 & 87.10 & 1.87 & ACS \\
 \hline
Wikitext & 64 & 1296 & 1314 & 82.78 & 86.83 & 4.05 & ACS \\
 & 96 & 1280 & 1314 & 81.46 & 85.67 & 4.21 & ACS \\
 & 128 & 1250 & 1301 & 77.97 & 79.82 & 1.85 & ACS \\
 & 160 & 845 & 889 & 69.66 & 80.53 & 10.87 & ACS \\
 & 192 & 529 & 564 & 81.50 & 75.45 & -6.05 & CS \\
 \hline
BookCorpus & 64 & 1907 & 1947 & 84.22 & 87.04 & 2.82 & ACS \\
 & 96 & 1873 & 1947 & 87.82 & 83.66 & -4.16 & CS \\
 & 128 & 1657 & 1749 & 84.74 & 85.49 & 0.75 & ACS \\
 & 160 & 863 & 922 & 83.59 & 83.68 & 0.09 & ACS \\
 & 192 & 476 & 518 & 79.43 & 83.38 & 3.95 & ACS \\
\bottomrule
\end{tabular}
}
\caption{Illustration of selective inference in aligning decoding strategies for language models. MAUVE scores are computed on both machine- and human-written text. In order to compute the KL-divergence, see Appendix~\ref{app:empirical-metrics}, MAUVE typically considers samples based on the minimal token lengths observed in human-reference texts (already seen during model training). Texts exceeding this length are truncated, while texts that do not achieve this length are excluded from the sample. Results are reported across three different datasets. This can be seen in the varying numbers of examples (column~3). Positive $\Delta$-values indicate that the decoding method ACS \cite{garces-arias-etal-2024-adaptive} outperforms the baseline CS \cite{su2023contrastivesearchneedneural}. All computations were performed using the \texttt{gpt2-xl} model \cite{radford2019language}. For CS \cite{su2023contrastivesearchneedneural}, hyperparameters $k = 5$ and $\alpha = 0.6$ were selected.}

\label{tab:mauve_inconsistencies}
\end{table*}

In this section, we discuss a tangible and concrete example of selective inference (Section~\ref{sec:stats-biases}) arising in empirical alignment of decoding methods in language models.  

When empirically aligning decoding methods via MAUVE, see also section~\ref{sec:decoding}, only machine-generated samples that match the distributional length constraints of human-written texts are considered. Any generated text that exceeds these bounds is often truncated or outright discarded \cite{pillutla2021mauvemeasuringgapneural,garces-arias-etal-2024-adaptive}. This leads to a biased evaluation where only text that “fits” pre-existing norms is measured, disregarding potentially superior but unconventional outputs. More precisely, MAUVE is typically computed over samples based on the minimal token lengths observed in human-reference texts (already seen during model training), in order to compute the Kullback-Leibler-divergence between their respective distributions, see Appendix~\ref{app:empirical-metrics}. Texts exceeding this length are truncated, while texts that do not achieve this length are excluded from the sample.

Through a statistical lens, any inference from a so-aligned model is doomed to suffer from post-selection bias -- as discussed in Section~\ref{sec:stats-biases}. Only considering samples of a certain length corresponds to reducing the domain of the sample, i.e, setting some parameters $\hat{\theta}$ to zero, thus shrinking the model's \textit{effective} parameter space $\Theta$. All subsequent probabilistic inference guarantees are then hampered, because they fail to condition on the events (that is, sample realizations) that led to the changes of $\Theta$. 
Correction methods (e.g., \citet{berk2013valid,benjamini1995controlling,wasserman2009high,fithian2014optimal,tibshirani2016exact,lee2016exact}) can render post-selection inference valid again. These prescriptive elements, however, are not considered in many empirical alignment use cases like decoding alignment through MAUVE \cite{pillutla2021mauvemeasuringgapneural}.

This highlights a general pattern: Forward empirical alignment, if driven by selective inference, may optimize for artifacts of the training data rather than the intended alignment objective. A naïve adherence to empirical alignment thus risks codifying and amplifying these biases in AI behavior rather than mitigating them.
By recognizing selective inference as a fundamental issue in empirical AI alignment, we advocate for more prescriptive and transparent approaches that explicitly account for sample selection biases and ensure alignment is guided by rationally defined objectives rather than self-reinforcing observational artifacts.

\section{Additional Alternative View: Empirical Alignment as a Necessary Benchmarking Tool}\label{app:alt-view}

We have argued against forward empirical alignment as a primary strategy. An alternative perspective holds that empirical methods are not just unavoidable but necessary \textit{for benchmarking and validation}. According to this view, even if prescriptive alignment is theoretically preferable, it ultimately requires empirical validation to ensure it functions as intended in real-world settings. Without empirical feedback, alignment objectives risk becoming too abstract or detached from practical deployment concerns \cite{gao2022aligning, eval_bias_2021}.

However, this argument presumes that empirical validation provides a neutral or reliable ground truth for alignment, rather than a reflection of existing biases. In practice, empirical benchmarks often reinforce anthropocentric limitations, locking models into human-like behaviors that may not generalize to more robust or scalable alignment strategies, see section~\ref{sec:stats-biases} and also \citet{gabriel2020artificial, perdomo2020performative,wang2024towards}. Instead of treating empirical evaluation as a fundamental necessity, an alternative approach would prioritize validation through formal guarantees, logical consistency, and predictive stability, reducing dependence on imperfect human-labeled data.
This perspective suggests to walk a tightrope, balancing empirical and prescriptive benchmarking: Rather than outright rejecting empirical methods, alignment research should redefine their role. One can treat them as a secondary check on prescriptive models rather than as the primary mechanism for defining alignment itself. 

In this framing, the burden of proof is reversed: Instead of requiring prescriptive methods to justify themselves empirically, empirical methods should justify their necessity in cases where prescriptive approaches provide clear, rule-based guarantees. This would allow empirical tools to serve as a complement rather than a crutch, making sure alignment strategies are both rigorous and adaptable.

\newpage
\bibliography{example_paper}

\bibliographystyle{icml2025}



\end{document}